\documentclass[letterpaper, 10 pt, conference]{ieeeconf}
\usepackage[pdftex]{graphicx}
\usepackage{cite}

\usepackage{amsmath}
\usepackage{amssymb}
\usepackage{amsfonts}			

\DeclareMathOperator*{\argminA}{arg\,min} 

\usepackage{eqparbox}

\usepackage{subcaption}			

\usepackage{color,soul}

\usepackage{tabularx}
\usepackage{xcolor}
\usepackage{multicol}
\usepackage{multirow}		
\usepackage{hhline}			

\title{\LARGE \bf
Closed-form Solution for IMU based LSD-SLAM Point Cloud Conversion into the Scaled 3D World Environment
}

\author{Sergey Triputen, Kristiaan Schreve, Viktor Tkachev and Matthias R\"atsch}

\begin{document}




\maketitle
\thispagestyle{empty}
\pagestyle{empty}

\begin{abstract}
SLAM is a very popular research stream in computer vision and robotics nowadays. For more effective SLAM implementation it is necessary to have reliable information about the environment, also the data should be aligned and scaled according to the real world coordinate system. Monocular SLAM research is an attractive sub-stream, because of the low equipment cost, size and weight. 

In this paper we present a way to build a conversion from LSD-SLAM coordinate space to the real world coordinates using a true metric scale with IMU sensor data implementation. 
The causes of differences between the real and calculated spaces are explained and the possibility of conversions between the spaces is proved.

Additionally, a closed-form solution for inter space transformation calculation is presented.
The synthetic method of generating high level accurate and well controlled input data for the LSD-SLAM algorithm is presented. 

Finally, the reconstructed 3D environment representation is delivered as an output of the implemented conversion.

\end{abstract}


\section{Introduction}

Robotic navigation and control is a complex task to deal with. Most positioning, orientation and odometry systems provide only pose data relative to the system start point. Therefore, it is necessary to accurately define the initial position and scale before motion starts. 
This is vital in order to integrate the SLAM system with other robot systems that interact with the environment. Several SLAM (Simultaneous Localization and Mapping) algorithms have been developed for this task. The main task of these algorithms is to build a map (based on the key-frames with loop closure detection) and also to propose a way to identify the current location of the robot during the localization phase \cite{Mur_Artal2014,Mur_Artal2015,Engel2014}.

Typical visual SLAM systems use either monocular, stereo or RGB-D cameras as a primary input device. Currently, systems using RGB-D cameras are most effective because they provide the depth data (distance to the detected point). The map can be used to calculate the scale factor of the camera path. However, relative to monocular systems, the hardware cost is the main disadvantage. Similar results can be achieved by combining monocular camera data with other sensors \cite{Usenko2016_1,Engel2015,Engel2014_2}.

Another important robot control task is the correct detection of environmental obstacles. 
Detailed environmental information is necessary for solving path planning tasks. 
In this paper a solution using a monocular SLAM system for 3D environment reconstruction is proposed, as well as map generation and further processing into metric scale is shown.
\begin{figure}[!t]
\centering

\begin{subfigure}{.5\textwidth}
  \centering
  \includegraphics[width=.9\linewidth]{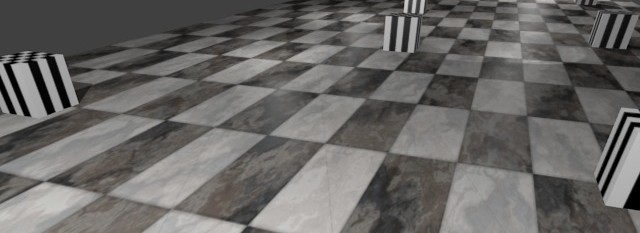}
  \caption{Blender environment modeled view.  This is used as an input for the frame sequence of the LSD-SLAM algorithm.}
  \label{fig:blenderEnv}
\end{subfigure}

\begin{subfigure}{.5\textwidth}
  \centering
  \includegraphics[width=\linewidth]{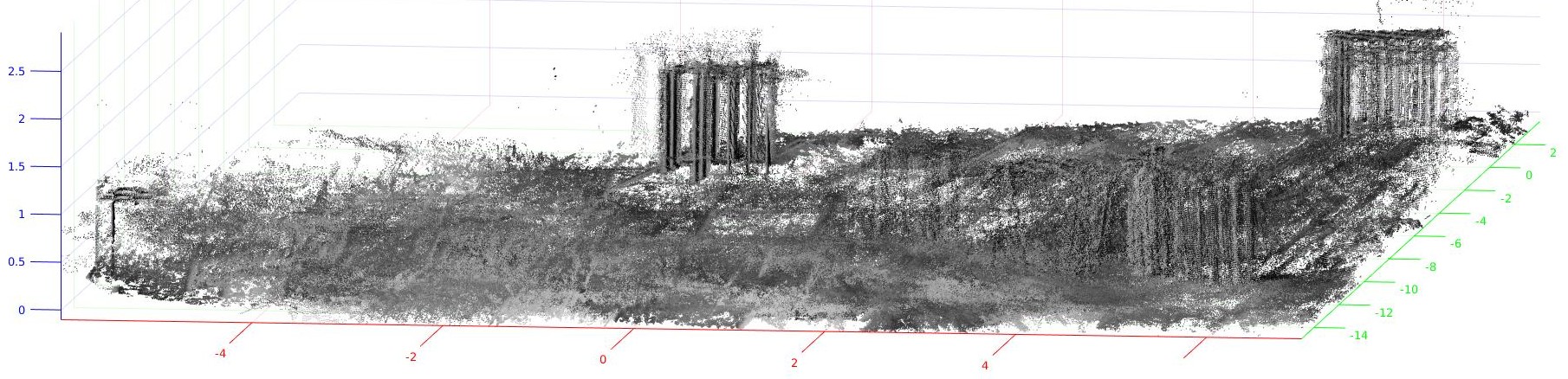}
  \caption{The visualization of the point cloud, that is already scaled to the real World metrics.}
  \label{fig:scaledEnv}
\end{subfigure}
\begin{subfigure}{.5\textwidth}
  \centering
  \includegraphics[width=\linewidth]{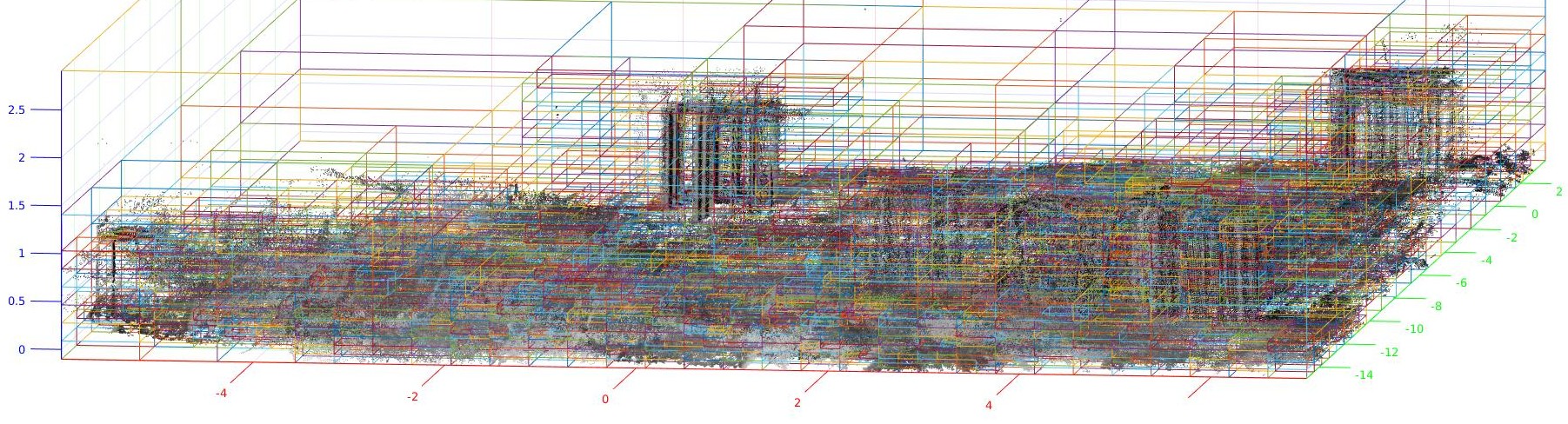}
  \caption{ Same geometry in MATLAB representation. The same point cloud with the octree representation} 
  \label{fig:octEnv}
\end{subfigure}

\caption{The results of conversion of point cloud LSD-SLAM algorithm estimated from blender frames into the real environment. Scaled according to the Earth surface.}
\label{fig:ENVs}
\end{figure}
There are many existing SLAM algorithms described today. 
Most of them are dealing with standard navigation tasks (e.g. building key-frame based maps, loop closure detecting or current position localization). 
One of the more popular approaches is PTAM \cite{Klein2007,Klein2009} and its variations, such as LSD-SLAM \cite{Engel2014,Engel2013} and ORB-SLAM \cite{Mur_Artal2014,Mur_Artal2015}.

PTAM and some other SLAM systems use feature point detection algorithms for frame tracking and finding the current camera position. 
These algorithms are computationally expensive and require significant resources. 
However some SLAM systems such as described in \cite{Montemerlo2002,Klein2009} are using a featureless solution and are optimized for embedded system implementation.

A significant challenge for monocular SLAM systems is the problem of estimating scaled camera positions. Without scaled metric data, the SLAM output cannot be integrated with other robotic systems relying on accurate environmental information. That is why some approaches use RGB-D or stereo cameras \cite{Endres2016,Maier2014}. Depth maps generated from RGB-D cameras can be used to find scaled metric camera positions. Additionally RGB-D hardware makes 3D environment reconstruction possible \cite{Endres2016}. This approach is more expensive than monocular systems.

However, scaled environmental information can be obtained if the monocular camera data is combined with other sensors, e.g. laser scanners, etc. The PTAM algorithm uses an altimeter based solution \cite{Engel2012}. PTAM does not provide scaled 3D environmental data, though.

There is no monocular SLAM system that provides a scaled point cloud representation of the environment aligned to some global coordinate system.

The main contribution of the paper is to present a method to combine LSD-SLAM estimated camera position with IMU sensor data in a real world coordinate system (which is perpendicular to the earth's surface). 
It is also shown how the aligned (position, orientation and scale) result can be used for octree mesh representation. 

\section{Spaces alignment problem solution}
A method to convert the estimated LSD-SLAM point cloud scale and orientation into the real world scale and absolute orientation is proposed in this paper. 
For the analysis of the results' accuracy we use simulation data, which is obtained from a synthetic world. It takes into account noise measurements of a real IMU sensor.

This chapter presents an overview of the problem and gives the necessary data annotation. 
Further, the problem solution proposal and the results evaluation criteria are described.

\subsection{Problem definition}\label{sub:problemDefination}
The main steps of LSD-SLAM and real world scale and orientation synchronization are taken during the LSD-SLAM algorithm initialization period. We should remember, that LSD-SLAM systems without spacial equipment (such as an external IMU sensor) have no information about the real world camera position, orientation and surrounding object sizes.

The LSD-SLAM camera position and orientation estimating process is based on tracking special points in a particular frame. It is necessary to have depth information for them. Initially, it uses random numbers for the depth hypotheses, i.e. $d\in[0,\,1]$ random numbers for each feature point in the first frame.

Also, for the first frame camera position and orientation the initial position vector  $[0,\ 0,\ 0]$ and quaternion $[1,\ 0, \ 0,\ 0]$ are used.

This random distance is used for camera tracking and for defining the estimated camera position and orientation. At the next step the calculated camera position and orientation error is used in the triangulation task to update the point cloud depth information. After some iterations, the calculation results are converged into an optimal combination of depths and camera tracking data.

It is obvious that after the initialization period the estimated point depths and camera position and orientation data have unpredictable values. The initialization time and final distance scale factor depend on a random distribution of initial hypothetical depths.

Now the task can be formulated: we are looking for a combination of rotation, scaling and translation factors that will map LSD-SLAM estimated coordinate system space to the IMU sensor coordinate system. 

It is very important to have the transformation from LSD-SLAM space to the sensor space, because the IMU sensor coordinate system is aligned to the gravity vector. That means it is aligned to the world coordinate system. It makes it possible to use an octree optimized 3D model representation for the estimated point cloud.  

\subsection{Problem solution}
In order to find a solution, first of all, it is necessary to formally define the input data. 
LSD-SLAM estimates data as a set of key-frames which describe camera position and orientation with a set of points or coordinates (or estimated distance from camera to each feature point). This will be denoted as $\mathbf{L}$.
\begin{equation}\label{equ:slamSet}
\mathbf{L}=\{T, \ \mathbf{P} \}; 
\end{equation}
Where $T\in\mathbb{S}\mathfrak{im}(3)$ such that:
\begin{subequations}\label{equ:sim3Diesc}
\begin{align}
T = \left( 
	\begin{array}{cc}
		s\mathbf{R} & \mathbf{t}	\\
		0  & 1
	\end{array}
	\right)
      \  	\textrm{with} 	\  		
 	&		\mathbf{R}	\in	\textrm{SO}(3) \label{equ:bigR},	\\ 
	& \ 	\mathbf{t} 	\in	\mathbb{R}^{3},						\\
	& \ 	s 			\in	\mathbb{R}^{+}		
\end{align}
\end{subequations}

The $\mathbb{S}\mathfrak{im}(3)$ group represents transformation in a three dimensional space.
There $s$ is a scale factor and $\mathbf{t}$  a translation vector in 3D space.

In the equation (\ref{equ:bigR}), $\mathbf{R}$ is an element of the $\textrm{SO}(3)$ group that represents rotation.
Each element of the $\textrm{SO} (3)$ group can be represented in several different forms (e.g. rotation matrix, quaternion or vector and angle combination).

 $T$ represents the camera transformation estimated by LSD-SLAM. Furthermore
\begin{subequations}\label{eqn:kfSet}
\begin{align}
	\mathbf{P} = \{ \ \mathbf{p} = \langle \ u,  v,  d \ \rangle  \ | &	\ u 	\in \mathbb{Z}^{+}, \\ 
    																&	\ v 	\in \mathbb{Z}^{+}, \\ 
                                                                	&   \ d		\in \mathbb{R}^{+} \ \};
\end{align}                                           
\end{subequations}
where $[u,\ v]$ are the pixel coordinates on the frame and $d$ is a distance from the camera position to the point.

In order to calculate 3D coordinates of the points in a key-frame point cloud, it is also necessary to have the intrinsic camera matrix:

\begin{equation}\label{eqn:cameraMatrix}
	M = \left[\begin{array}{c c c}
	f_{x} 	& 		0  	&  c_{x}	\\
	0	 	&	f_{y} 	&  c_{y}  	\\
	0		& 		0  	&  1  		\\
	\end{array}\right]
\end{equation}

Where:

$\mathbf{f}=\left[ f_{x}, \ f_{y} \right]$ - camera focus distance per axis.

$\mathbf{c}=\left[ c_{x}, \ c_{y} \right]$ - image position of the principle axis in pixel coordinates. 

Let $\mathbf{X}=\{ \ \mathbf{x} \ | \ \mathbf{x} \in \mathbb{R}^3 \ \}$  represent $\mathbf{P}$ in the 3D space of the LSD-SLAM coordinate system. It can be calculated by mapping:
\[		
	\delta \colon \mathbf{P}\rightarrow \mathbf{X}		
\] 
Where $\delta$ was specified like:
\begin{equation}\label{eqn:pointCorrdiantesCalculation}
\mathbf{X}=\delta( \ \mathbf{L}, M \ ) = T \left[ 
\begin{array}{c}
\frac{\left( u-c_{x}\right) \cdot d}{f_{x}}		\\
\frac{\left( v-c_{y}\right) \cdot d}{f_{y}}		\\
d												\\
1
\end{array}
\right]	
\end{equation}

We are receiving additional information from an external sensor, in this case an IMU. Similar to equation (\ref{equ:slamSet}) for LSD-SLAM estimated data, we will use set $\mathbf{S}$, which describes sensor measured data  
\[ \mathbf{S}=\{T^{'}, \ \mathbf{P}^{'} \}; \]

For IMU sensor measured data $T^{'}\in\mathbb{S}\mathfrak{im}(3)$ is a transformation into the real world metric coordinate system and $\mathbf{P}^{'}\equiv\varnothing$ is an empty set, because the IMU model do not have any information about the 3D real world geometry. Because $\mathbf{P}^{'}$ is empty, $\mathbf{X}^{'}$ cannot be obtained.

As described in the previous subsection, we are looking for a combination of rotation, translation and scale conversions from $\mathbf{L}$ to $\mathbf{S}$. 

Formally that conversion combination can be represented as an element of $\mathbb{S}\mathfrak{im}(3)$:

%
\begin{align*}
\Lambda = \left( 
	\begin{array}{cc}
		s\mathbf{R} & \mathbf{t}	\\
		0  & 1
	\end{array}
	\right)
	\  	\textrm{with} 	\  		
	&	\mathbf{R}	\in	\textrm{SO}(3),			\\
	& \ \mathbf{t} 	\in	\mathbb{R}^{3},		\\
	& \ 	s 		\in	\mathbb{R}^{+}		
\end{align*}

Finally, we can formalize our task as follows: we are looking for a mapping
\[		
	\Lambda \colon \mathbf{L}\rightarrow \mathbf{S};		
\] 

In order to find $\Lambda$ we can only use the LSD-SLAM camera position and orientation and then relate it to the IMU sensor measurements. 
As already mentioned $\mathbf{P^{'}}$ (hence $\mathbf{X^{'}}$ as well) is an empty set for the sensor coordinate system. 
Therefor the equation for finding $\Lambda$ becomes:

\begin{equation}\label{eqn:labdaCalculation}
\Lambda^{*} := \argminA_{(s,\mathbf{R},\mathbf{t})} 
\left( 
\sum \left\Vert  \ T^{'} - \Lambda \times T \ \right\Vert^{2} 
\right) 
\end{equation}

The element usage is based on the fact that  $T$ and  $T^{'}$ are part of the $\mathbb{S}\mathfrak{im}(3)$.
The solution to this equation is discussed in detail in \cite{Horn1987, Horn1988}.
	
And so, we are able to reconstruct the point cloud in the IMU  sensor coordinate system.
With the calculated $\Lambda^{*}$, we can transform the LSD-SLAM estimated point cloud to the IMU sensor coordinate system. This can be achieved using equation \ref{eqn:pointCorrdiantesCalculation} and the result of equation \ref{eqn:labdaCalculation}.

\begin{equation}\label{eq:pointCloudCalculation}
\mathbf{X^{'}}	\approx	\Lambda^{*}	\times \mathbf{X} 
				\approx \Lambda^{*}	\times \delta( \ \mathbf{L}, M \ ) 
\end{equation}

\subsection{Closed-form solution}
Horn \cite{Horn1987} gives the closed-form solution for solving equation \ref{eqn:labdaCalculation}. This solution is summarized in this algorithm:

\begin{enumerate}
\item Find the centroids $\mathbf{t}$ and $\mathbf{t^{'}}$ of the LSD-SLAM estimated $\mathbf{T}$ and 
IMU sensor measured $\mathbf{T^{'}}$ sets:

\begin{align*}
& \mathbf{\bar{t} }	= \frac{1}{n} \sum_{i=1}^{n}\mathbf{t }_{i} \\
& \mathbf{\bar{t}}'	= \frac{1}{n} \sum_{i=1}^{n}\mathbf{t'}_{i}
\end{align*}

where $n$ is a frames number.
\item The centroids are subtracted from all elements of sets

\begin{align*}
&\mathbf{\hat{t} }	= \mathbf{t } - \mathbf{\bar{t }} \\
&\mathbf{\hat{t}}'  = \mathbf{t'} -\mathbf{\bar{t'}}
\end{align*}

so that from now on, we deal only with values relative to the centroids. 

\item For each pair of coordinates we compute the nine possible products
$\hat{t_{x}}\hat{t_{x}'}, \ \hat{t_{x}}\hat{t_{y}'},..., \ \hat{t_{z}}\hat{t_{z}'}$ of the components of the two vectors.
This is added up to obtain $S_{xx}, \ S_{xy},..., \ S_{zz}$, where

\begin{equation}
S_{xx} = \sum_{i=1}^{n}\mathbf{\hat{t_{x}} }\mathbf{\hat{t_{x}}}',  \quad
S_{xy} = \sum_{i=1}^{n}\mathbf{\hat{t_{x}} }\mathbf{\hat{t_{y}}}'
\end{equation}

and so one.
These nine totals contain all the information that is required to find the solution.

\item The rotation quaternion finding $\mathbf{\dot{e}}$ by solving homogeneous equation
\[
		[N-\lambda I]\mathbf{\dot{e}}=0
\]
There matrix $N$ is determined as 
\begin{equation*}
N = \left[	\begin{array}{cccc}
			a & e & h & j	\\
			e & b & f & i	\\
			h & f & c & g	\\
			j & i & g & d	\\
			\end{array}
    \right] 
\end{equation*}
with elements
\begin{align*}
a& = ( \ \ S_{xx} + S_{yy} + S_{zz}), 	&b& = ( \ \ S_{xx} - S_{yy} - S_{zz}), 	\\
c& = (-S_{xx} + S_{yy} - S_{zz}), 		&d& = (-S_{xx} - S_{yy} + S_{zz}), 	\\ 
e& = ( S_{yz} - S_{zy} ), 				&f& = ( S_{xy} + S_{yx} ), 			\\
g& = ( S_{yz} + S_{zy} ), 				&h& = ( S_{zx} - S_{xz} ),			\\
i& = ( S_{zx} + S_{xz} ),				&j& = ( S_{xy} - S_{yx} )
\end{align*}
\item At this point, compute the scale as:
\[
	s = \frac{ \sum_{i=0}^{n} \mathbf{\hat{t}' \cdot R(\mathbf{\hat{t} }) }}
    		 { \sum_{i=0}^{n} \Vert \mathbf{\hat{t}' \Vert}^{2} }
\]

\item Computation of the translation as the difference between the centroid of the IMU sensor measurements and the scaled and rotated centroid of the LSD-SLAM estimated data
	\[
    	\mathbf{r_{0}} = \mathbf{\bar{t}}' - s\mathbf{R(\bar{t})}
    \]
where $\mathbf{r_{0}} \in \mathbb{R}^{3}$ is translation vector.

\end{enumerate}

\subsection*{Note:}\label{chapt:scaleFuctor}
The case, described in \cite{Horn1987} assumes the two sets which are well enough synchronized from the start point. In our case, there are several unsynchronized initial sets. It is recommended to exclude these unsynchronized sets. To solve this task we propose to analyze frame to frame vector length factors. There factors should be calculated as:

\begin{equation}\label{equ:scaleFuctor}
	p_{k} = \frac{ \Vert \mathbf{t}'_{k+1} - \mathbf{t}'_{k} \Vert^{2} }
    			 { \Vert \mathbf{t}_{k+1} - \mathbf{t}_{k} \Vert^{2}}
\end{equation}

where k is the frame number.

As a result we present a method to transform point cloud SLAM data into the IMU sensor coordinate system (gravity vector aligned). 


\section{Input data estimation}

All SLAM algorithms rely on and are very sensitive to input data and ambient conditions (such as color gradient, light, contrast etc.). 
In oder to evaluate the quality of the obtained conversion we need to have precise information of the camera position and orientation as an input. 
The best way to achieve precise camera position and orientation data is to make a controlled simulation with known parameters (including environment dimensions). 
An animation application tool which makes this possible is 'Blender'.
\begin{figure}[ht]
	\centering
	\includegraphics[width=\linewidth]{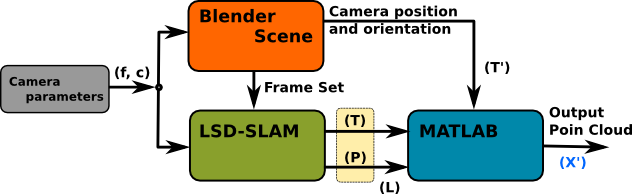}
	\caption{Structure scheme of conversion data flow of the initial camera frame into the scaled and aligned point cloud data.}
\label{fig:schema}
\end{figure}
%
\subsection{Basic scene overview}
\begin{figure}[ht]
\centering
	\includegraphics[width=.9\linewidth]{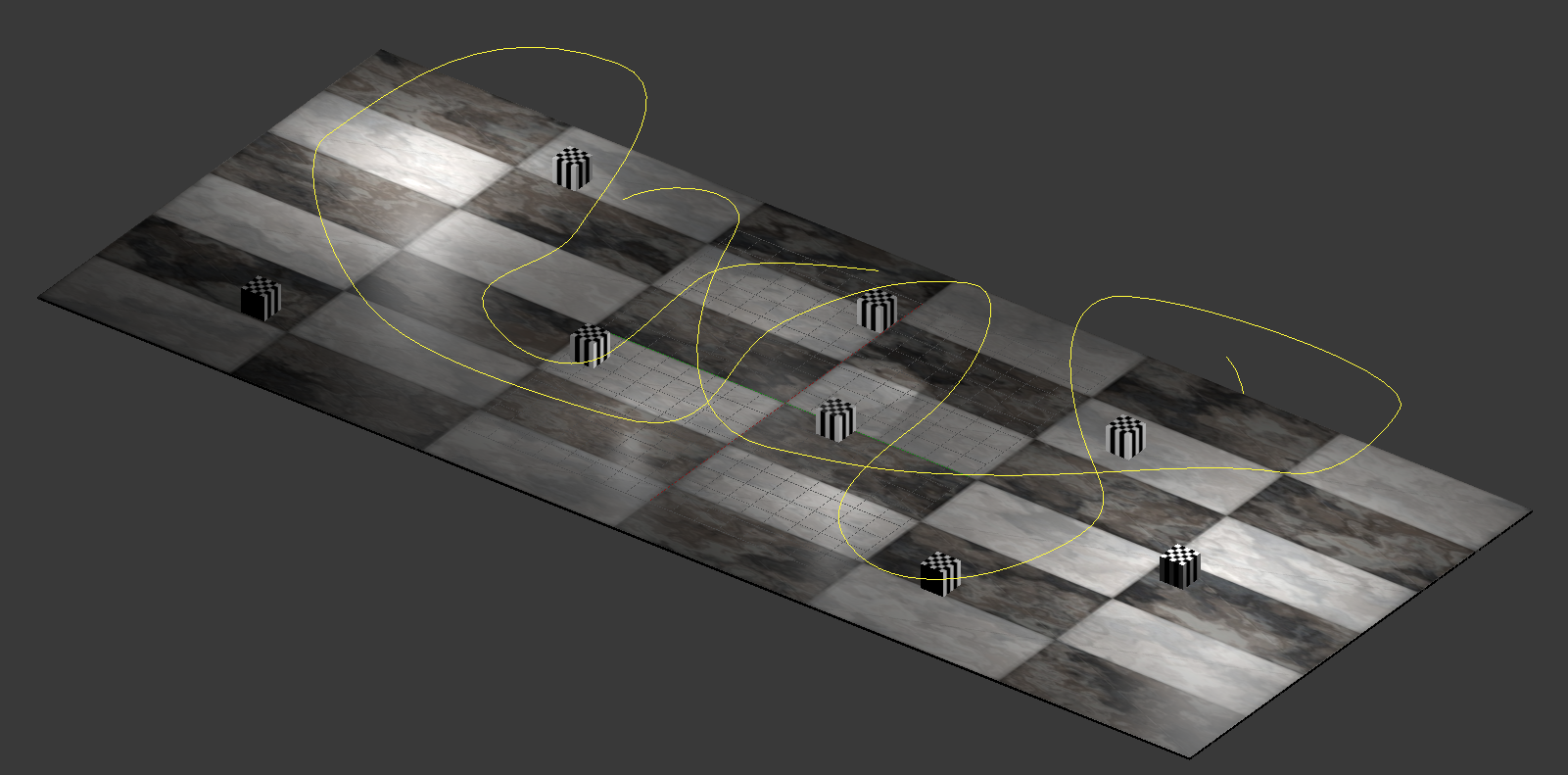}
	\caption{Test Blender scene for conversion research.}
\label{fig:blenderScene}
\end{figure}

For our experiments the scene represented on Fig. \ref{fig:blenderScene} is used. 
It is a $10\times5$ meters plane with eight $1\times1\times1$ meters cubes. 
This scene size is enough to finish the initializing process of LSD-SLAM and to receive a set of "clear" key-frames for point cloud analyses. 

\subsection{Camera parameters}

In order to have realistic results from LSD-SLAM, we need a definite set of camera parameters. We assume the parameters described below.
\begin{itemize}
\item Camera Frame Per Second (FPS). 
\\
FPS determines the achievable camera motion speed. The higher the FPS, the higher camera motion speed may be.
\item Frame Resolution.
\\
Selection of this parameter is a balance between having enough resolution to locate feature points and computational time.
\item Lens Focus Distance.
\\
This parameter is used to calculate the 3D position of the feature point, as shown in equations \ref{eqn:cameraMatrix} and \ref{eqn:pointCorrdiantesCalculation}.

\item Lens Distortion Parameters.
\\
For cameras with a big field of view ("Fish Eye" effect) It is very important to use distortion coefficients (camera undistorted). For our experiments we model a non "Fish eyed" camera, so the mentioned coefficients are equal to zero.

\end{itemize}

For the simulation we have used camera parameters of the Sony PlayStation Eye 3 as an input to LSD-SLAM.

PS Eye 3 was selected since tests in our laboratory have demonstrated its success with LSD-SLAM.

The camera parameters are stated in Table \ref{tab:Camera}.

\begin{table}[ht]
\caption{PS3 EYE CAMERA SPECIFICATIONS}  
\centering 						
\begin{tabular}{l|l} 		    
\hline\hline 					
Resolutions						& $640\times480$ pixels @ 60  Hz 	\\
\hline
Field Of View  					& $75^\circ$			\\ 
\hline\hline
\end{tabular} 
\label{tab:Camera} 
\end{table}  

\subsection{External environment parameters}

\begin{itemize}

\item Geometry dimensions

For the present work it is very important to have a method to evaluate point cloud accuracy. It is not a trivial task in real word applications. If tested in real environments, it will be very complicated to compare the point cloud and camera position data with, e.g. 3D scanner or sensor data.

We used the Blender application to build a virtual world with well-known geometric dimensions and to export it into MATLAB for final analysis.

\item Flexible texture control and light exposition.

In Blender, unlike real world camera perception, it is possible to control the LSD-SLAM inputs (such as light, shade, contrast, etc.). 
This makes it possible to create precise environmental model scenarios to verify the algorithm results. See Fig. \ref{fig:Shadow} for an example of a typical scene used in this study.
\begin{figure}[t]
\centering
\begin{subfigure}[t]{.24\textwidth}
  \includegraphics[width=0.92\linewidth]{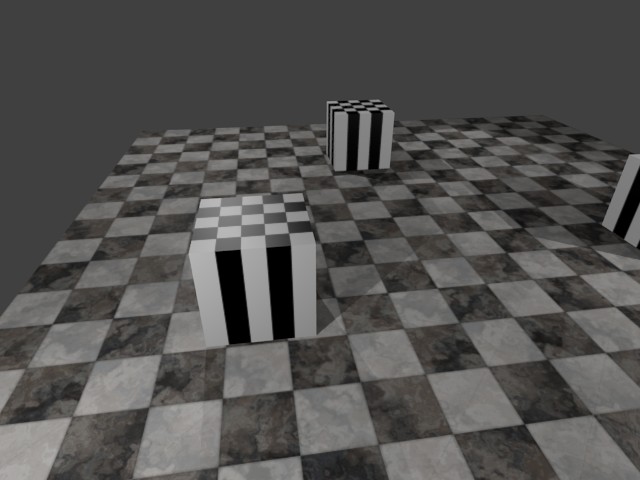}
  \caption{Frame with contrast textures ans shadow edges.}
  \label{fig:Shadow_origin}
\end{subfigure}\hfill
\begin{subfigure}[t]{.24\textwidth}
  \includegraphics[width=\linewidth]{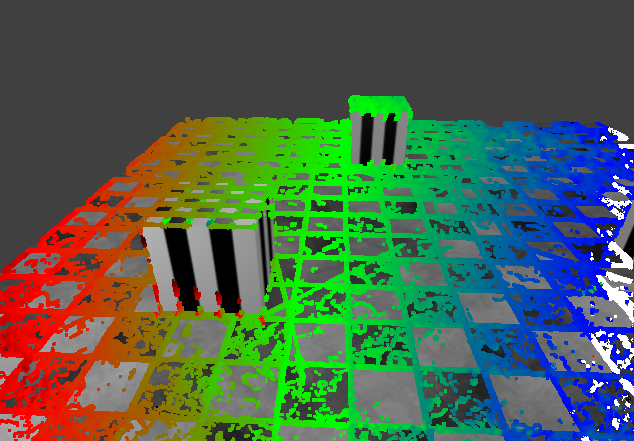}
  \caption{Selected and processed by LSD-SLAM algorithm.}
  \label{fig:Shadow_depth}
\end{subfigure}

\caption{Using virtual environment Blender reproduces real inputs (e. g. shadow) for LSD-SLAM, generating quasi-real results.}
\label{fig:Shadow}
\end{figure}
\end{itemize}
%
%
\subsection{Camera position and orientation (IMU or odometry data simulation)}

Similar to environmental input for the LSD-SLAM algorithm analysis, it is a beneficial to have a controlled simulation of camera position and motion.

\begin{itemize}

\item Camera trajectory (length, loops, curvature angels)

From the beginning of the simulation we have a specified 'zero' position. Then, the ground truth path is controlled, taking into account the IMU data.

The simulation ensures no motion occurs which makes LSD-SLAM results unpredictable (e.g. camera angular velocity, uncertainty of feature points due to insignificant rotation angle, distance, etc.).

\item Camera motion speed (constant or variable along path) is also simulated in a controlled way to allow accurate interpretation of LSD-SLAM results.
\\

\end{itemize}

	In order to eliminate real environment, robot and camera input inaccuracy, simulation with Blender is proposed. The simulation technology provides enough spacial dimension and environment resolution to recreate real environment parameters (e. g. light, shade, gradient).

	LSD-SLAM algorithm results with simulated input are tested and delivered with accurate results if compared to the algorithm results with the real camera and environment input data.



\section{Results}
\begin{figure}[t]
    	\centering
		\includegraphics[width=.9\linewidth]{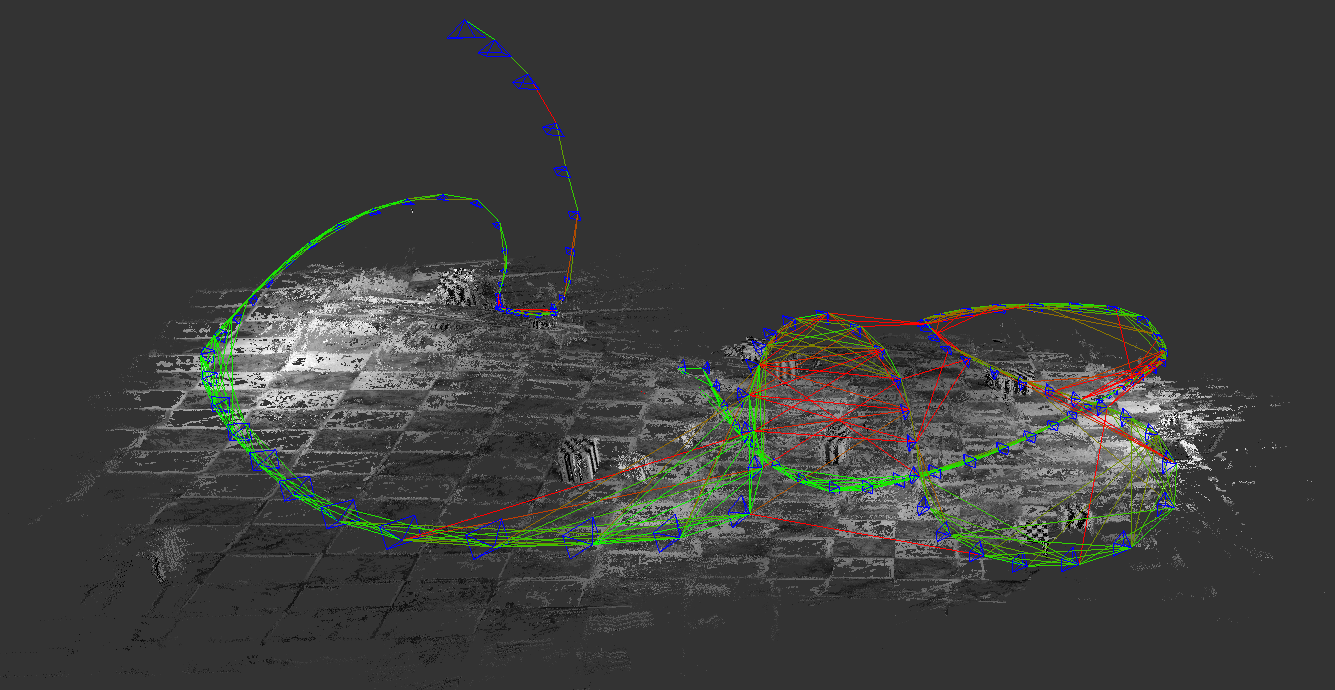}
		\caption{ Point cloud and camera path estimated by an original LSD-SLAM project and LSD-SLAM viewer represented}
		\label{fig:originalPath}
\end{figure}

The set of frames taken of the Blender scene (Fig.\ref{fig:blenderScene}) is used an as input for LSD-SLAM.
The results are received from LSD-SLAM and are represented by the LSD-SLAM viewer as shown in Fig.\ref{fig:originalPath}.

The point cloud in this figure orientation was chosen manually in order to make the presentation more understandable.

\subsection{Basic scene analyzing}

\begin{figure}[hb]
\centering
	\begin{subfigure}{.5\textwidth}
	\centering
		\includegraphics[width=.85\linewidth]{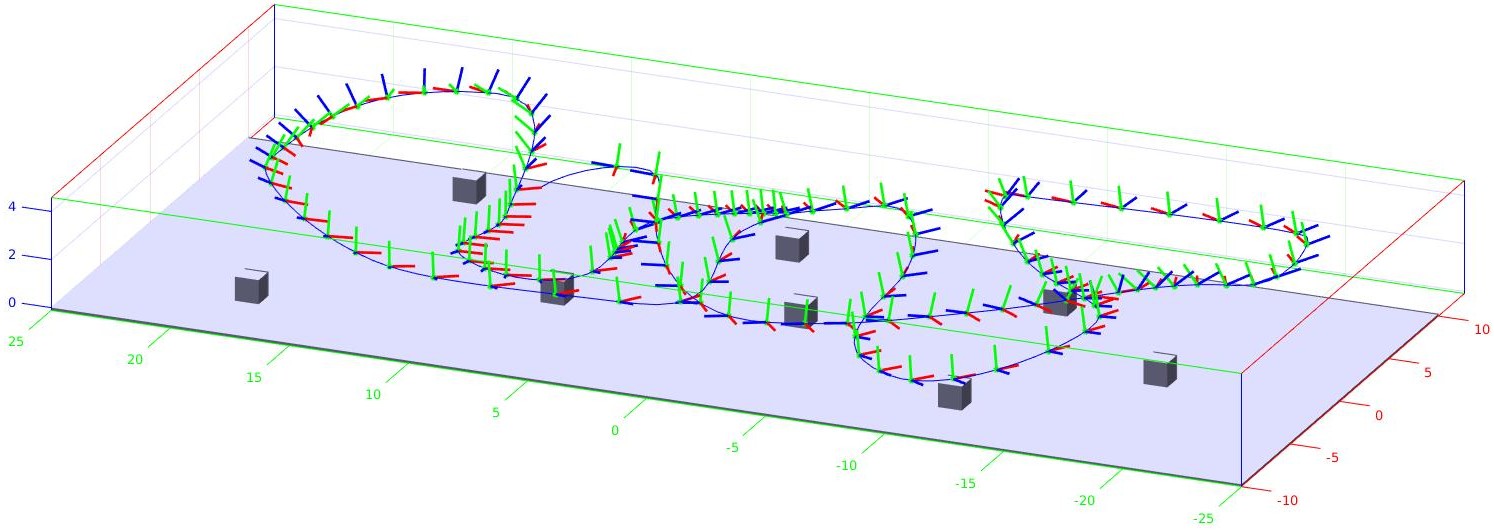}
		\caption{Camera path exported from Blender }
		\label{fig:blenderPath}
	\end{subfigure}
	
	\begin{subfigure}{.5\textwidth}
	\centering
		\includegraphics[width=.85\linewidth]{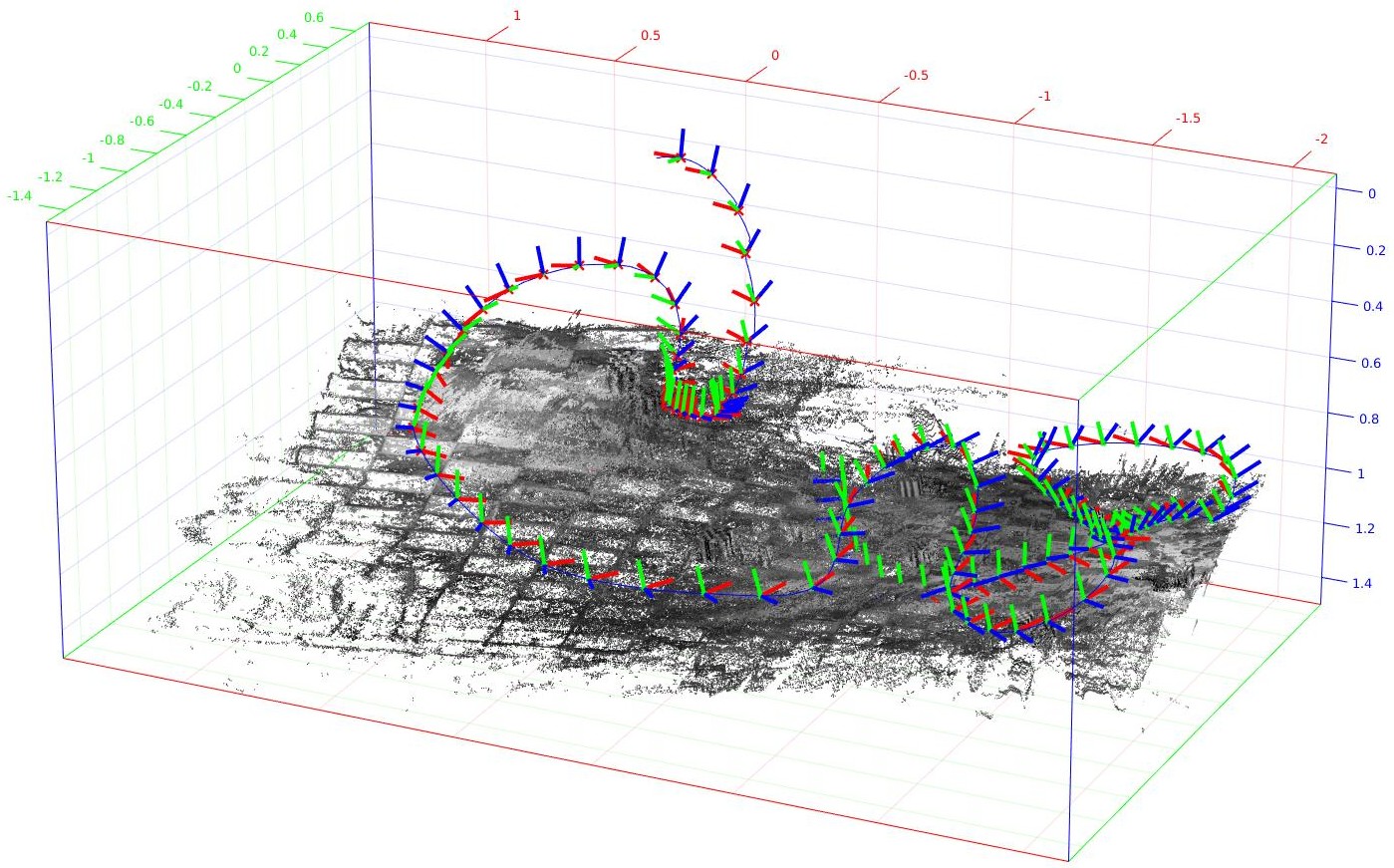}
  		\caption{Point cloud and camera position and orientation estimated by LSD-SLAM and presented in MATLAB axises }
  		\label{fig:slamPath}
	\end{subfigure}
\caption{Ground truth and LSD-SLAM camera path comparison.}
\label{fig:originalPathes}
\end{figure}

In order to make the comparison of the size and orientation of point cloud and camera path, MATLAB is used (Fig.\ref{fig:slamPath}).

The ground truth path and scene geometry were exported from Blender and imported to MATLAB. It is represented in (Fig.\ref{fig:blenderPath}).

As expected, there is a big difference in the scale factor and orientation representation between the two figures. 
It is possible to see this by comparing the axis sizes.

Additionally, \ref{fig:blenderPath} and \ref{fig:slamPath} show that XYZ axes (presented with Red, Green and Blue respectively) do not match by orientation.

The basic scene parameters are:
\begin{itemize}
\item Total frames number - 7000;
\item Key frames number	 - 123;
\item Simulation period - 116.7 seconds;
\item Blender path length 	 - 195.12 ground truth meters;
\item LSD-SLAM estimated path length 	 - 12.6 LSD-SLAM internal units.
\end{itemize}

\subsection{Initialization time analyzes}

\begin{figure}[ht]
\centering
  \includegraphics[width=\linewidth]{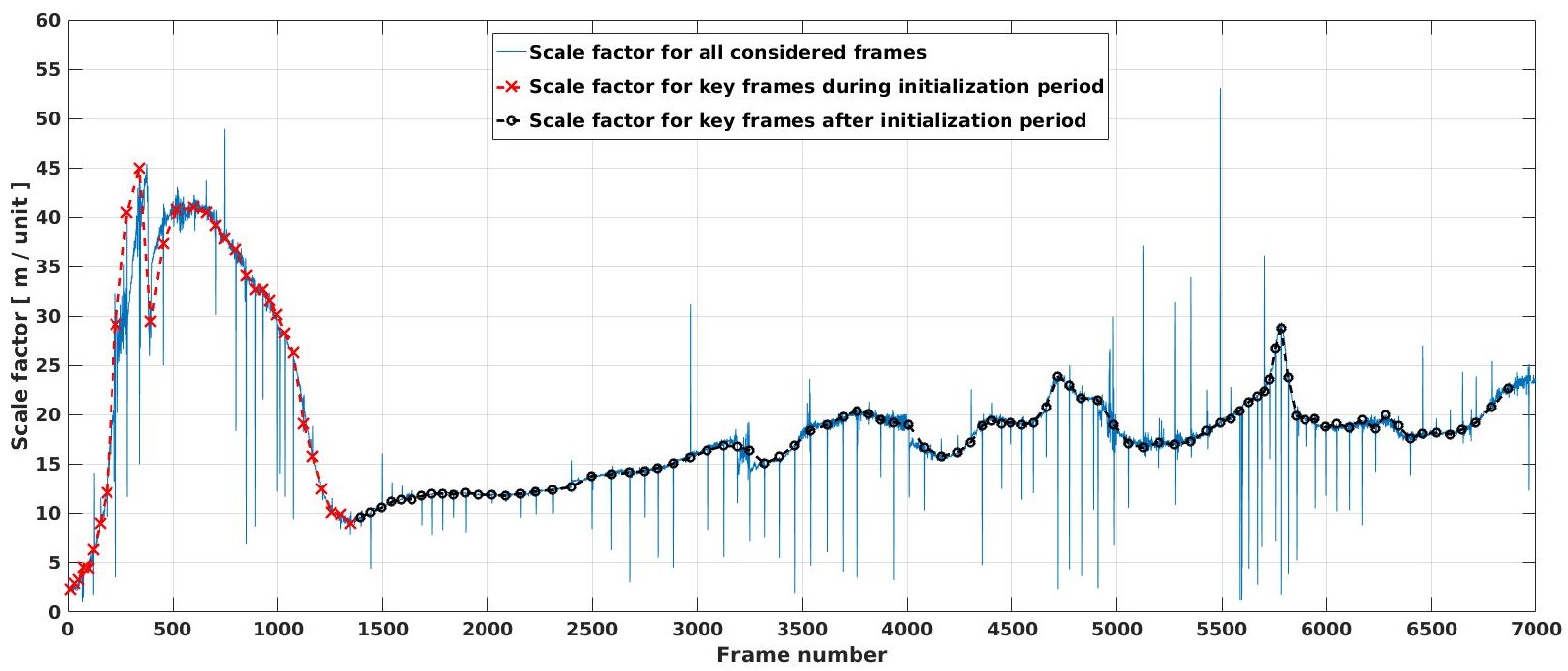}
\caption{Scale factors analyzing.}
\label{fig:scaleFactor}
\end{figure}

Now we can try to analyze the initialization time. 
We used equation \ref{equ:scaleFuctor} from chapter \ref{chapt:scaleFuctor}. 
The calculation results are represented in Fig.\ref{fig:scaleFactor}.

The depicted trend is showing the three key stages:

1.	Obviously occasional deviations (the starting stage, shown in red dots).

2.	Convergence period (the second stage with outlined trend, shown in black dots, $1100-4300$ counts). Still not stabilized.

3.	Stabilized stage, after 4300 count. Still has visible bursts, but trend is kept stable (black dots after 4300).

%
\subsection{Conversion results}
\begin{figure}[ht]
\centering
\begin{subfigure}[t!]{.5\textwidth}
  \centering
  \includegraphics[width=.95\linewidth]{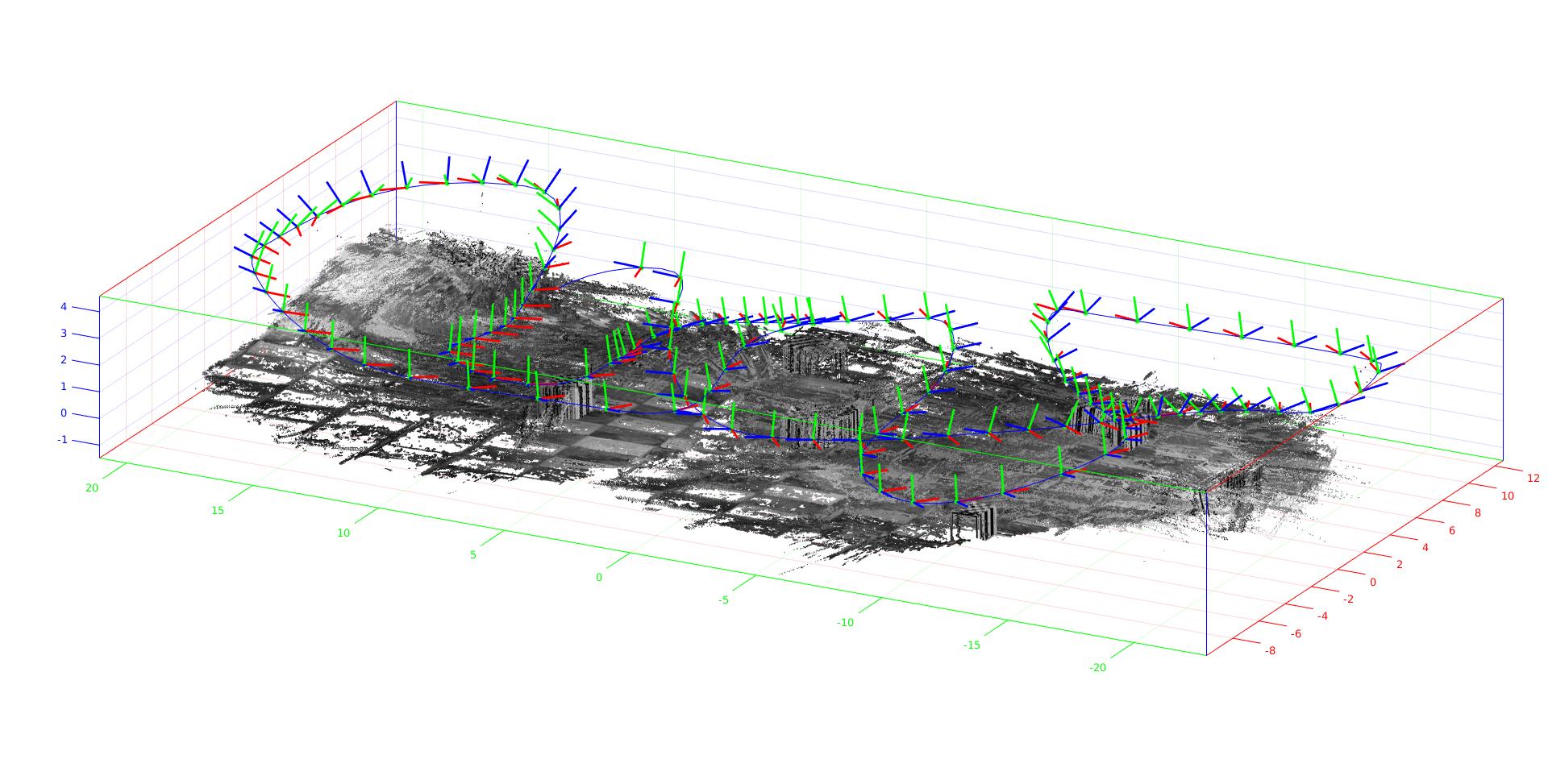}
  \caption{Full scene after conversion applied.}
  \label{fig:convertedPointCloudFull}
\end{subfigure}
\begin{subfigure}[t!]{.5\textwidth}
  \centering
  \includegraphics[width=.95\linewidth]{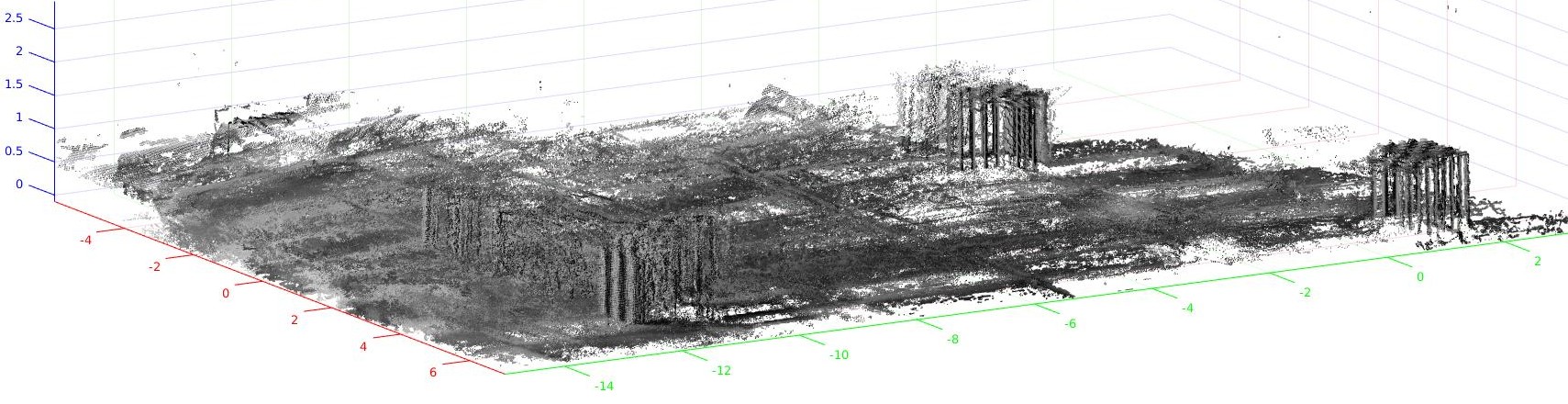}
  \caption{A fragment of the converted scene.}
  \label{fig:convertedPointCloudPart}
\end{subfigure}
\caption{Conversion results.}
\label{fig:convertedResults}
\end{figure}

The conversion results of the position and orientation of the camera and the point cloud coordinates are represented in Fig.\ref{fig:convertedResults}. 

Fig.\ref{fig:convertedPointCloudFull} shows that the obtained point cloud coordinates are aligned according to the Blender. As well as the axis directions and scales are aligned.

Fig.\ref{fig:convertedPointCloudPart} shows a fragment of the converted scene.

%



\section*{CONCLUSION}
According to the achieved results, the targeted transformation shows that the precision for estimated 3D environment for octree map representation is enough according to the objective of this article. 

    It is important to mention that LSD-SLAM algorithm gives high precision for the point cloud. Consequently the obtained 3D reconstruction is reliable to be used in “path planning” or “follow me” tasks. 
    
At the same time, there are several technical points revealed in the current research that are to be considered.
One of the mentioned technical points: the scale coefficient is not constant all the time. According to the motion path, if a frame is repeated frequently, the scale should be estimated and adjusted in the process.

\bibliographystyle{IEEEtran}
\bibliography{IEEEabrv,biblio_rectifier}


\end{document}